\documentclass[runningheads]{llncs}
\usepackage{graphicx}
\usepackage{subcaption}
\usepackage{color}
\usepackage{multirow}
\usepackage{hyperref}

\begin{document}
\title{Self-supervised pre-training enhances change detection in Sentinel-2 imagery}

\author{Marrit Leenstra\inst{1} \and
Diego Marcos\inst{1}\orcidID{0000-0001-5607-4445} \and
Francesca Bovolo\inst{2}\orcidID{0000-0003-3104-7656} and Devis Tuia\inst{1,3}\orcidID{0000-0003-0374-2459}}
\authorrunning{M. Leenstra et al.}
\titlerunning{PRRS @ICPR 2020/2021 preprint}

\institute{Wageningen University, the Netherlands.\email{diego.marcos@wur.nl} \and
Fondazione Bruno Kessler, Trento, Italy. \email{bovolo@fbk.it} \and Ecole Polytechnique Fédérale de Lausanne, Sion, Switzerland. \email{devis.tuia@epfl.ch}
}
\maketitle              

\begin{abstract}
\textbf{This is the preprint version of the paper published in the Pattern Recognition and Remote Sensing workshop (PRRS) 2021, held in ICPR 2020/2021 (virtual). For the postprint, please refer to the official LNCS publication of the conference proceedings.\\}
While annotated images for change detection using satellite imagery are scarce and costly to obtain, there is a wealth of unlabeled images being generated every day. In order to leverage these data to learn an image representation more adequate for change detection, we explore methods that exploit the temporal consistency of Sentinel-2 times series to obtain a usable self-supervised learning signal. For this, we build and make publicly available {(\url{https://zenodo.org/record/4280482})} the Sentinel-2 Multitemporal Cities Pairs (S2MTCP) dataset, containing multitemporal image pairs from 1520 urban areas worldwide. We test the results of multiple self-supervised learning methods for pre-training models for change detection and apply it on a public change detection dataset made of Sentinel-2 image pairs (OSCD).

\end{abstract}
\section{Introduction}

Large amounts of remote sensing images are produced daily from airborne and spaceborne sensors and can be used to monitor the state of our planet. Among the last generation sensors, the European Copernicus program has launched a series of satellites with multispectral sensors named Sentinel-2 (S2 hereafter). S2 has a revisit time between five days (at the Equator) and 2-3 days at mid-latitudes. With such high revisit rate, change detection, \emph{i.e.} the comparison of images acquired over the same geographical area at different times to identify changes \cite{Liu:2019:review_multispectral_cd}, {allows for near real-time monitoring of dynamics that are observable though remote sensing, including} forest monitoring  \cite{Verbesselt:2010:trend_seasonal_timeseries,Hamunyela:2016:spatial_context_bfast}, urbanisation mapping \cite{Deng:2009:urbanization,Huang:2017:urbanisation} and disaster monitoring \cite{Brunner:2010:multi_sensor_CD_disaster,Longbotham:2012:data_fusion_contest_cd}.

Many change detection methods have been proposed in the literature~\cite{Bovolo:2015:time_cd}. They tend to identify changes either by comparing classification maps~\cite{Vol10e} or by first extracting some kind of index to be thresholded to highlight changes \cite{Vol14b}. Recently, deep learning has been considered to learn how to align data spaces, so that changes are better highlighted and easier to detect~\cite{Lin:2019:multispectral_bilinearCNN,Zhan:2017:siamese_cd,Peng:2019:UNet++,Mou:2019:CNN+RNN,Saha:2019:deepCVA}.

Despite the success of these approaches, the {lack of} a relevant and large labeled dataset {limits their applicability~\cite{Zhu:2017:DL_remote_sensing}.}
{In computer vision tasks using natural images, it is common to use models that have been pre-trained on a large dataset for a loosely related task. A different number of bands and image structure limits the usability of these models to S2 imagery. This exacerbates the need for a tailored change detection ground truth,} which is often difficult to obtain: especially when change is a rare anomaly (\emph{e.g.} after a disaster), there are no labeled sets to train deep learning models on.

To decrease the amount of supervision, one can revert to models using types of annotation requiring less human effort. One could use exploit the geometry of data manifolds by using semi-supervised models, or change the type of annotations, for example by considering weak supervision, \emph{e.g.} image-level annotations rather than pixel level ones~\cite{Kel19d} or imprecise labels~\cite{daudt2019gad}. These approaches are successful, but still require some level of supervision provided by an annotator.

In this paper, we explore the possibility of reducing this requirement to a minimum. We consider strategies based on \emph{self-supervised learning}~\cite{Doersch:2015:self-supervised_spatial_context,Caron:2018:self-supervised_clustering}, where a neural network is trained using labels extracted directly from the images themselves. Rather than training the model on the change detection task, we train it on a \emph{pretext task} for which the labels can be extracted from the image pairs directly (\emph{e.g.} relative locations of patches). By doing so, we can pre-train the majority of the weights and then teach the model to recognize changes with a minimal amount of labels. We create a large and global dataset of S2 image pairs, S2MTCP, where we train our self-supervised learning model, before then fine-tuning it on the OSCD change detection dataset~\cite{Daudt:2018:OSCD_CD_dataset} for pixel-level change detection. The results show that achieving state of art change detection is possible with such a model pre-trained without labels, opening interesting perspectives on the usage of self-supervised learning in change detection.

\section{Methods}
In this section, we present our entire pipeline (Section~\ref{sec:overall}) and then detail the self-supervised pretext tasks used for pre-training (Section~\ref{sec:ssl}).

\subsection{Change detection pipeline}\label{sec:overall}
Let $I^1$ and  $I^2$ be two multispectral images acquired over the same geographical area at time $t^1$ and $t^2$ respectively. 
We want to pre-train a model on a set of unlabeled images $\{U = (I_u^1, I_u^2)_i\}_{i=1}^{N}${ such that it can be} easily fine-tuned on a {small} set of labeled image pairs $\{L = (I_c^1, I_c^2)_i\}_{i=1}^{M}$.

\begin{figure}
\includegraphics[width=\textwidth]{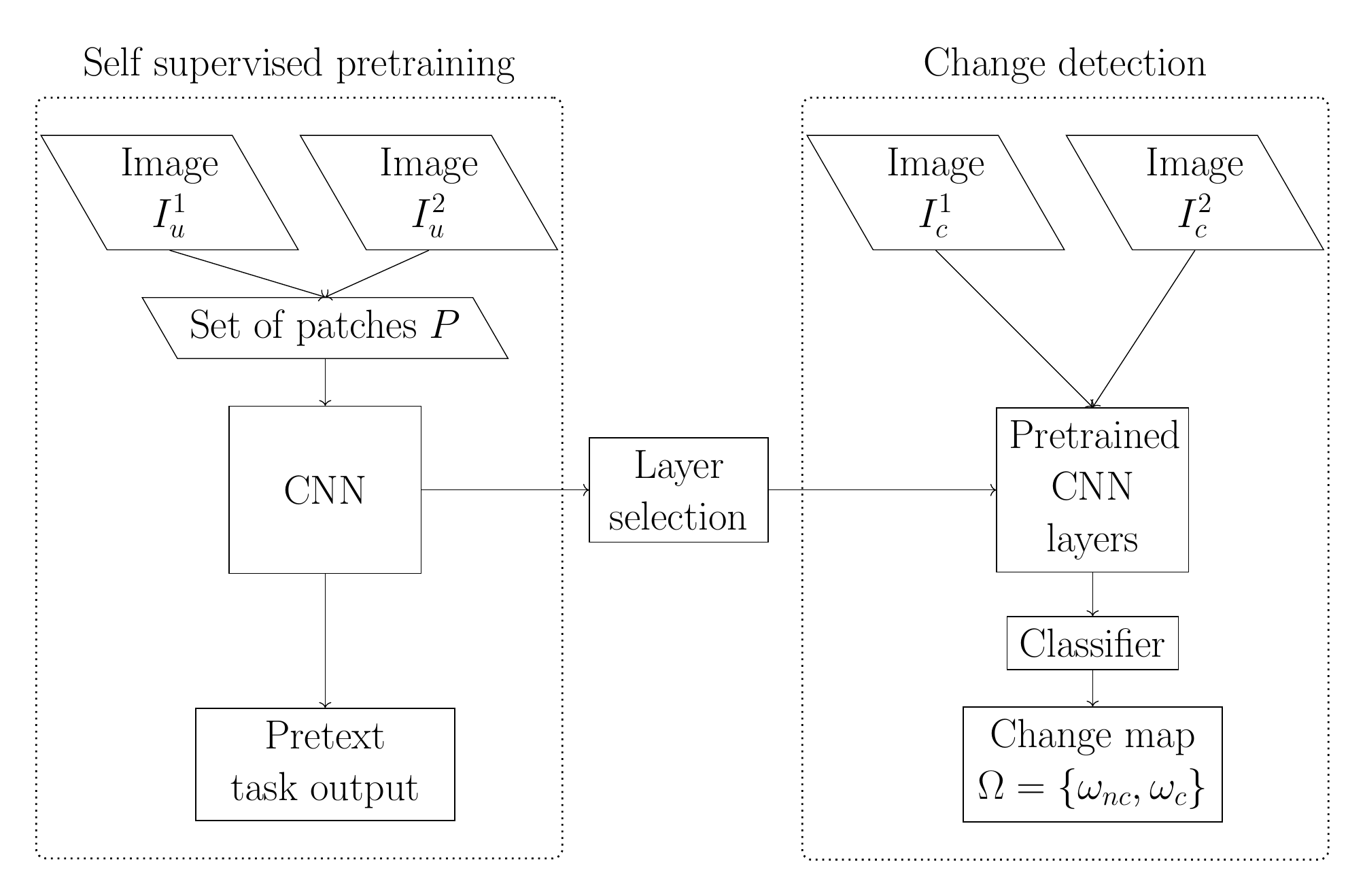}
\caption{Overview of the methodology.} \label{fig:methodology}
\end{figure}

The overall pipeline comprises three phases: first the network is trained on the pretext task (see Section~\ref{sec:ssl}), then the layer with the best features for change detection is manually selected. Finally, these features are used in a second network performing  change detection. Figure \ref{fig:methodology} presents the overview of the methodology. 

\subsubsection{Phase 1: self-supervised pre-training.} Ideally, we would like the change detection network to be able to focus on learning the changed areas. To do so, one would hope that the low level features in the change detection network \emph{align} the two image radiometric spaces, so that
the {features} for $I_c^1$ and $I_c^2$ become similar for areas were no changes have occurred. 

To facilitate this process, we learn such features using a self-supervised task on a large, unlabeled dataset, $U$. {This task has to be related to the task of change detection so that the learned features become useful.}
We test two different pretext tasks:  (1) discriminate between overlapping and non-overlapping patches and (2) minimizing the  difference between overlapping patches in  feature space. Both pretext tasks are described in detail in the next Section \ref{sec:pretext_tasks}.

\subsubsection{Phase 2: feature selection.} 
The deeper layers in the network are likely to be more task-specific, which means that earlier layers might be more suitable for the downstream task \cite{Gidaris:2018:self-supervised_rotation}. Therefore, we add a feature layer selection step to extract the feature layer that results in the highest change detection performance. Image pairs $(I_c^1, I_c^2)_i$ are passed as input to the network and, at each layer the activation features $\mathbf{f}_{l,i}^1$ and $\mathbf{f}_{l,i}^2$ are extracted. {A linear classifier is} then trained on top of features extracted from a specific layer $l$. 
Performance of each layer is manually compared, and the layer with the highest performance is selected for the change detection task. 

\subsubsection{Phase 3: change detection.} 
The selected layer is used to extract features from the change detection image pairs. We discriminate between unchanged ($\omega_{nc}$) and changed ($\omega_c$) pixels, based on the assumption that the unchanged pixels result in similar features and the changed pixels yield dissimilar features. Two classifiers are compared for this task: (1) a linear classifier and (2) Change vector analysis ({CVA}, \cite{Bovolo:2015:time_cd}). The linear classifier is trained in a supervised way on the complete training set $L$, by minimizing the weighted cross entropy loss. {CVA} is an unsupervised method and does not require any training. However, note that the classification with {CVA} is not fully unsupervised as {at this stage} ground reference maps were used to select the optimal feature layer. {However, solutions can be designed to make the selection procedure unsupervised.}  \newline \newline

\subsection{Pretext tasks for self-supervision}\label{sec:ssl} \label{sec:pretext_tasks}
{In self-supervised learning, a pretext task is an auxiliary learning objective on which the model is pre-trained. Although not identical to the final task (self-supervised learning is there to pre-train models when there are not enough labels for the final task), this auxiliary objective is designed such that it helps the model learn features that are expected to be useful on the final task.}

Several pretext tasks have been proposed in self-supervised learning literature{: for example, \cite{Doersch:2015:self-supervised_spatial_context}  predicts relative positions of nearby patches, while \cite{Gidaris:2018:self-supervised_rotation} rotates patches and predicts such rotation for enforcing invariances. Regardless of the specific implementations,} the common  denominators are that (1) the pretext labels must be extracted from the images themselves without external supervision and (2) the pretext task must help learn features that are relevant for the real downstream task (in our case detecting changes). In the previous section we discussed the need of the change detection network to learn features that project unchanged pixels pairs in the same part of the feature space (\emph{i.e.} unchanged areas become more similar \cite{Vol14b}).
To learn features in this direction, we propose two pretext tasks: 

\begin{figure}[!t]
\centering
\begin{subfigure}{0.9\textwidth}
\includegraphics[width=0.95\linewidth]{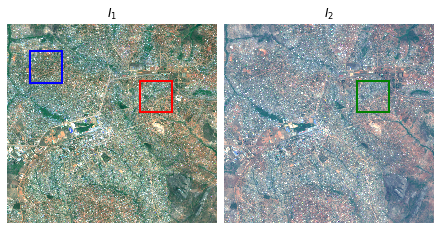}
\caption{Location of patches in the images pair.}
\label{subfig:pretext_tasks_a}
\end{subfigure} 
\hfill

\begin{subfigure}{0.45\textwidth}
\includegraphics[width=0.95\linewidth]{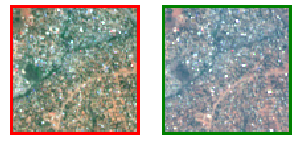} 
\caption{Overlapping patch pair}
\label{subfig:pretext_tasks_b}
\end{subfigure}
\begin{subfigure}{0.45\textwidth}
\includegraphics[width=0.95\linewidth]{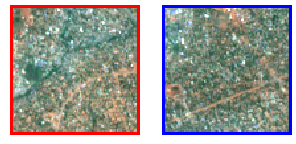}
\caption{Non-overlapping patch pair}
\label{subfig:pretext_tasks_c}
\end{subfigure}

\begin{subfigure}{0.67\textwidth}
\includegraphics[width=0.95\linewidth]{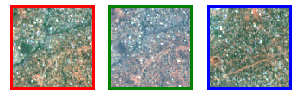}
\caption{Patch triplet}
\label{subfig:pretext_tasks_d}
\end{subfigure} 

\caption[Illustration of patch sampling strategy.]{Illustration of the patch sampling strategy for the self-supervised learning tasks. (a) Patches are spatially and temporally randomly sampled in the unlabelled image pair $(I_u^1, I_u^2)$. The colored squares represent the  patches locations. (b) Overlapping patch pair (red and green) for pretext Task 1. The associated pseudo label $y_{j} = 0$. (c) Non-overlapping patch pair (red and blue) for pretext Task 1. The associated pseudo label $y_{j} = 1$. (d) Patch triplet for pretext Task 2. 
} 
\label{fig:pretext_tasks}
\end{figure}

\begin{enumerate}
\item The first pretext task is defined by a binary classification that requires the network to predict whether or not a patch pair is overlapping. Each training example $\mathit{P_j}$ contains a patch pair $\{(\mathit{p^1}, \mathit{p^2})_j,y_j\}$. The associated pseudo label equals $y_{j} = 0$ for spatially overlapping pairs and $y_{j} = 1$ for spatially non-overlapping ones. The patch pairs are spatially and temporally randomly sampled from the unlabelled image pairs, and equally divided over the two classes. The task is illustrated in Figure \ref{subfig:pretext_tasks_a}-\ref{subfig:pretext_tasks_c}.

The underlying hypothesis is that sampling $\mathit{p^1}$ and $\mathit{p^2}$ randomly from either $I_u^1$ or $I_u^2$ learns the model to ignore irrelevant radiometric variations due to acquisition conditions and to focus on relevant spatial similarity/dissimilarity between patches. 
The parameters of the network are optimized by minimizing binary cross-entropy loss, given by 
\begin{equation} \label{eq:loss_binary_cross_entroply}
    L = - (y_{j} \cdot log({P}(y_{j})) + (1-y_{j}) \cdot log(1 - {P}(y_{j}))) 
\end{equation}
where ${P}(y_j)$ is the probability of pseudo label $y_j$ given input $P_j$ as calculated by the logistic sigmoid function in the output layer of the network. 

\item The second pretext task aims to learn image representations that project overlapping patches close to each other in the high dimensional feature space and non-overlapping patches far away. The patch sampling strategy is similar to the one of the first pretext task, with patches spatially and temporally randomly sampled in unlabelled image pairs. However, each training example $P_j$ contains one extra patch to form patch triplets $({p^1}, {p^2}, {p^3})_j$. Patches ${p^1}$ and ${p^2}$ are spatially overlapping, while ${p^3}$ is not (Figure \ref{subfig:pretext_tasks_a} and \ref{subfig:pretext_tasks_d}.). 

The distance between features extracted from overlapping patches ${p^1}$ and ${p^2}$ should be close to zero, while the distance between feature extracted from disjoint patches ${p^1}$ and ${p^3}$ should be larger by a margin $m$. This can be accomplished by minimizing the triplet margin loss with an additional $\ell_1$ loss. The complete loss function is given by
\begin{equation}
    L = max(||\mathbf{f}^1 - \mathbf{f}^2||_2 - ||\mathbf{f}^1 - \mathbf{f}^3||_2 + m, 0) + \gamma \cdot |\mathbf{f^1} - \mathbf{f^2}|
    \label{eq:pt2}
\end{equation}
where $\mathbf{f^i}$ is the feature vector for patch ${p^i}$ and $\gamma$ is a hyperparameter to balance the triplet loss and the $\ell_1$ loss functions. %
\end{enumerate}

\begin{figure}
    \centering
\begin{tabular}{cc}
    \includegraphics[height=5cm]{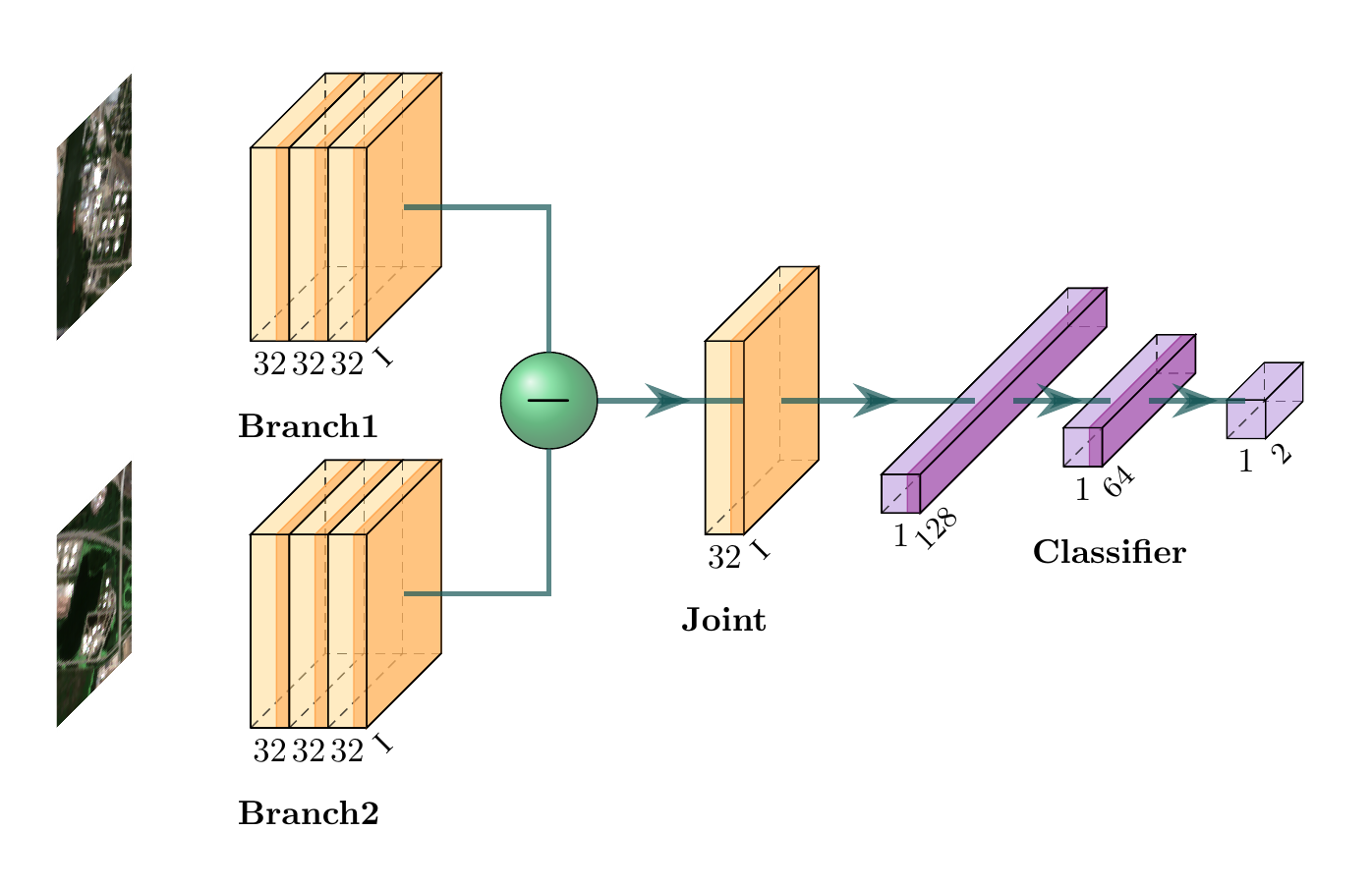} &
    \includegraphics[height=5cm]{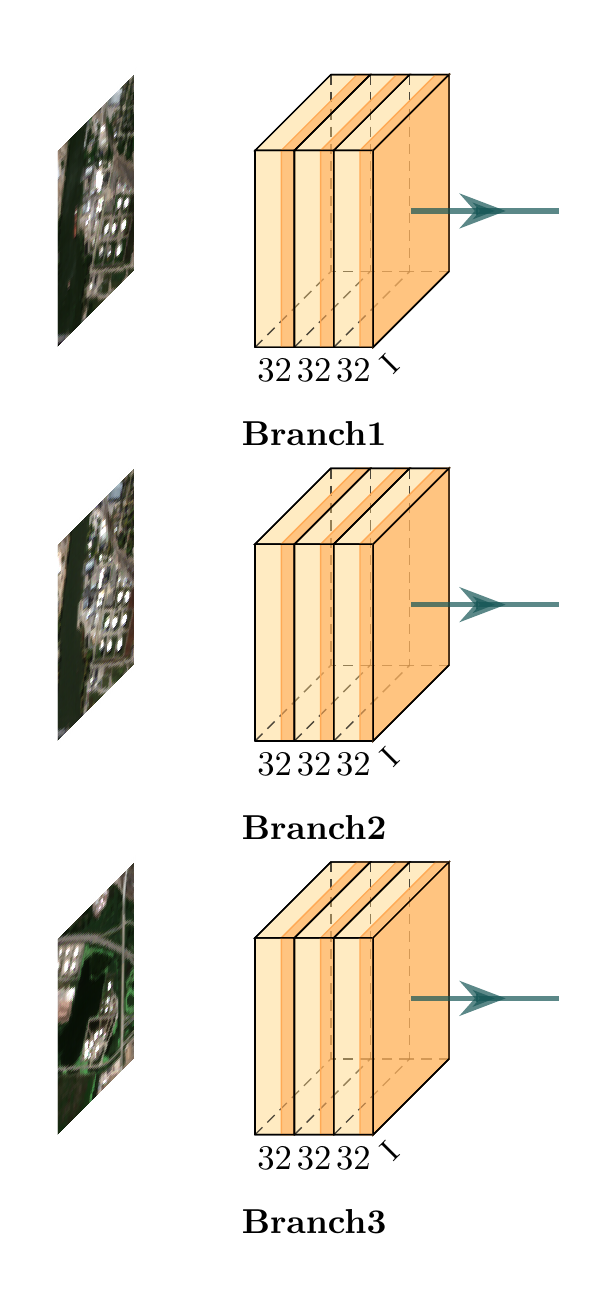} \\
    (a) Pretext Task 1 & (b) Pretext Task 2 \\
    \end{tabular}
    \caption{Schematics of the architecture of the self-supervised CNNs.}
    
    \label{fig:siamese}
\end{figure}

{The network for the first pretext tasks is implemented as a Siamese architecture with three convolutional layers per branch and a fusion layer, as shown in Fig.~\ref{fig:siamese}a, while the second one does not require the fusion layer, Fig.~\ref{fig:siamese}b.}

\section{Data and setup}

\subsection{Datasets}
\subsubsection{Change detection.}\label{sec:datal} For the change detection task, we use the OSCD benchmark dataset \cite{Daudt:2018:OSCD_CD_dataset} with annotated urban changes. It contains 24 S2 image pairs with dense reference labels $\{(I_c^1, I_c^2)_i, \Omega_i\}_{i=1}^{24}$ where $\Omega \in \{\omega_{nc}, \omega_c\}$. Images are approximately 600x600 pixels and contain scenes with different levels of urbanisation. The dataset is originally divided into 14 labeled pairs with freely available ground reference maps. The labels of the remaining  10 test pairs are only available through the DASE data portal (http://dase.grss-ieee.org/) for independent validation. In this work, 12 images are used as training set; we used the two remaining images to evaluate the change maps qualitatively. Quantitative results in the discussion section are computed on the 10 undisclosed images, after upload of the obtained maps to the DASE data portal. 

\subsubsection{Sentinel-2 multitemporal cities pairs (S2MTCP) dataset} 
A dataset of  S2 level 1C image pairs $U =\{(I_u^1, I_u^2)_i\}_{i=1}^{N}$, was created for self-supervised training. As the scope of this research is limited to urban change detection, the image pairs were focused on urban areas. Locations are selected based on two databases containing central coordinates of major cities in the world \cite{dataset_simplemaps,dataset_geonames} {with more} than 200.000 inhabitants. 

Image pairs $(I_u^1, I_u^2)_i$ are selected randomly from available S2 images of each location with less than one percent cloud cover. Bands with a spatial resolution smaller than 10 m are resampled to 10 m and images are cropped to approximately 600x600 pixels centered on the selected coordinates. Hence, every image covers approximately 3.6km$^2$. According to the Sentinel User Guide \cite{sentinel:2015:user_handbook}, level 1C processing includes spatial registration with sub-pixel accuracy. Therefore no image registration is performed. 

\begin{figure}[!t]
    \centering
    \includegraphics[width=\linewidth]{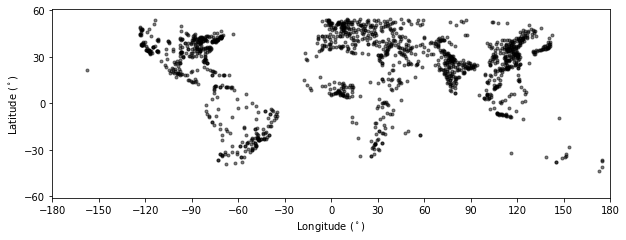}
    \caption{Location of the cities sampled in the generated S2MTCP dataset.}
    \label{fig:location_worldcities}
\end{figure}

The S2MTCP dataset contains $N=1520$ image pairs, spread over all inhabited continents, with the highest concentration of image pairs in North-America, Europe and Asia (Fig.~\ref{fig:location_worldcities}). The size of some images is smaller than 600x600 pixels. This is a result of the fact that some coordinates were located close to the edge of a Sentinel tile, the images were then cropped to the tile border. {It is available at the URL \url{https://zenodo.org/record/4280482}}.

\subsection{Setup}
\subsubsection{Self-supervised pretraining setup.} \label{sec:setup_pretext_task_performance}
We use 85\% of the S2MTCP dataset $U$ to train the model, and use 10\% to validate it. We keep the remaining 5\% as a blind test set for numerical evaluation.

The parameters are optimized using the Adam optimization algorithm \cite{kingma:2014:adam} with the suggested defaults for the hyperparameters ($\beta$1 = 0.9, $\beta$2= 0.999). The training is stopped when the validation loss does not decrease by  1\% in between epochs. We use a fixed learning rate of $0.001$ and weight decay ($0.0001$). The $\gamma$ parameter in Eq.~(\ref{eq:pt2}) is set to 1 experimentally. At each iteration, we sample 5 patch pairs (or triplets for pretext Task 2) from each image to generate {6350 patch pairs per epoch.} Data augmentation (90 degrees rotation and horizontal/vertical flips) are applied. 

\vspace{0.2cm}

To assess the performance on the pretext tasks, we use the blind test set extracted from $U$. For pretext Task 1, we assess the success rate in the task itself in percentage, while for Task 2, we consider the value of the loss. We also run the pretext tasks on the {12} images composing OSCD test set {to assess domain shifts}. Note that no OSCD labels are used at this stage.

\subsubsection{Feature layer selection setup} 
The performance of features $\mathbf{f}_l$ on the change detection task is compared using 3-fold cross validation on the OSCD labeled set. As discussed in Section~\ref{sec:datal}, the OSCD labeled set contains 12 image pairs ($(I_c^1, I_c^2), \Omega$), hence, we use 4 pairs per fold. We consider features (\emph{i.e.} activation maps) at different levels of the self-supervised model as candidates for the selection. In other words, we retain features $\mathbf{f}^{\{1,2\}}_l$, with $l=[1,...,3]$, where $l$ is the depth of the CNN considered (see schematics of Fig.~\ref{fig:siamese}) for images $I^1_c$ and $I^2_c$, respectively. We use the differences of the corresponding features as inputs for the change detection classifier. For pretext Task 1, we also consider $l=4$, \emph{i.e.} the substraction layer where $\mathbf{f}^1_3$ and $\mathbf{f}^2_3$ are fused.

{The linear classifiers} are trained for a maximum of 250 epochs and stopped if the validation loss does not improve for 50 epochs. The same optimizer and augmentation used in the previous step are used.
We sample 100 patches pairs per image of the OCSD dataset. 
To make sure that the results for each experiment (varying layer and pretext task) are comparable, the patches are passed to the classifiers in the same order. Performance is evaluated based on F1-score, sensitivity, specificity and precision. 

\subsubsection{Change detection setup} 
Two classifiers are compared for the change detection task: 
\begin{itemize}
    \item \textit{Supervised linear classifier}, trained in a supervised way on the OSCD training dataset.   {This model consists of a single linear layer followed by a sofmax activation function returning the probability scores $\{\omega_{c}, \omega_{nc}\}$. The threshold to obtain the change binary map was set based on the F1-score on the training set.}
    
     \item \textit{CVA} \cite{Bruzzone:2013:cd_framework}, with detection threshold optimised using either Otsu's method or the triangle method~\cite{Rosin:2001:triangle}.
      
\end{itemize}

The CV folds and extracted patches are the same as in the feature layer selection step. Same goes for optimization and augmentation strategies. The learning rate was decreased to $10^{-5}$. 
\section{Results and discussion}

\subsubsection{Pretext tasks performance} 
The validation and test results for pretext Task 1 (\emph{i.e.} predicting whether two patches are spatially overlapping) are reported in Table \ref{tab:results_pretexttask1}. The test accuracy was consistently high in both datasets: in all cases the model was able to correctly predict whether the patches were overlapping in over 97\% of the patch pairs. The low number of epochs required to reach this high accuracy indicates the pretext task was easy to solve. 
Regarding Task 2,  the lowest validation loss was reached after 17 epochs and training stopped. The loss on the OSCD dataset was slightly higher than on the S2MTCP dataset (result not shown), as a result of a larger contribution of the triplet loss. We argue that this does not indicate overfitting, but rather a domain gap between the two datasets, since the difference between the validation and test loss on the S2MTCP dataset remains small. 

\begin{table}
\caption{Performance on pretext Task 1, expressed in {Average} Accuracy (\%).}
\centering
\begin{tabular}{c|c|c|c} \hline
Dataset & Data split & Loss & Accuracy \\ \hline \hline
S2MTCP & validation & 0.043 & 98.93 \\
S2MTCP & test & 0.052 & 98.28 \\ 
OSCD & - & 0.083 & 97.67 \\
\hline
\end{tabular}
\label{tab:results_pretexttask1}
\end{table}

\subsubsection{Selection of optimal feature layer for change detection.} 
Table \ref{tab:results_AA_layer_selection} presents the average accuracy  over the three folds for change detection performed with features $\mathbf{f}_l$ for layers $l \in [1,4]$. The features of the second convolutional layer ($l=2$) perform best in both cases, although the differences are overall small. The performance of the deeper layers in the network trained on pretext task~1 decreases faster than the performance of the ones trained on pretext task~2. It is not surprising that features from deeper layers perform worse on the change detection task, Yosinski et al. \cite{Yosinski:2014:transferring_features} have shown that deeper layers of a CNN are specific to the task and dataset used for training, while the first layers are general-purpose. This effect has also been observed when transferring features from a pretext task to the target task in self-supervised learning \cite{Kolesnikov:2019:self_supervised_comparison}.  

Based on these results, the second convolutional layer is selected for the change detection task.
\begin{table}[!t]
\caption[Evaluation of features per layer as measured by Average Accuracy on the change detection task.]{Evaluation of features per layer as measured by Average Accuracy (\%) on the change detection task by cross validation. $l = [1,3]$ represents which convolutional layers of the self-supervised model is used. 
For each pretext task the best performance is highlighted in bold text.}  
\centering
\begin{tabular}{l|l|l|l|l} \hline
Pretext task &  $l=1$ & $l=2$ & $l=3$ & $l=4$ \\
\hline\hline
Pretext Task 1 & 76.06 & \textbf{77.82} & 75.92 & 74.26 \\
Pretext Task 2 & 78.03 & \textbf{79.11} & 78.19 & \\
\hline
\end{tabular}
\label{tab:results_AA_layer_selection}
\end{table}

\subsubsection{Numerical results on the OCSD test set.} As a final step, we compare the results of our self-supervised model with those obtained by fully supervised models on the undisclosed test set on the DASE algorithm testbed data portal (see section~\ref{sec:datal} for details). 

\begin{figure}[!t]
    \centering
    \begin{tabular}{cc|ccc}
    \multicolumn{1}{c}{} &&& Pretext Task 1 & Pretext Task 2\\
    \includegraphics[width=.30\linewidth]{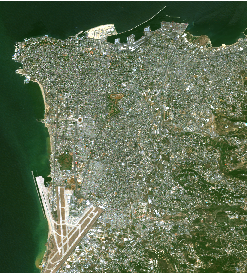} &&& \includegraphics[width=.30\linewidth]{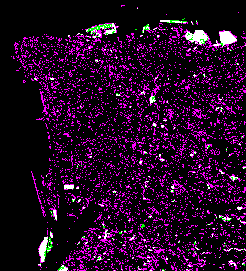} &\includegraphics[width=.30\linewidth]{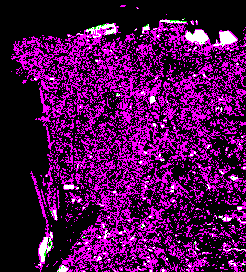}\\ 
    Image, $t_1$&&& \multicolumn{2}{c}{CVA, Otsu's method} \\
    \includegraphics[width=.30\linewidth]{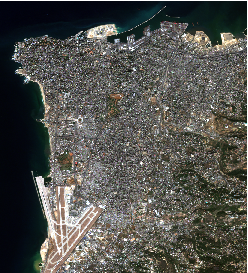} &&& \includegraphics[width=.30\linewidth]{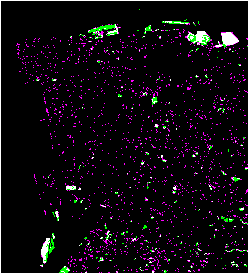} &\includegraphics[width=.30\linewidth]{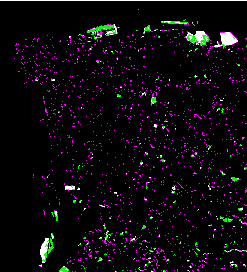}\\
    Image, $t_2$&&& \multicolumn{2}{c}{CVA, triangle method} \\
    \includegraphics[width=.30\linewidth]{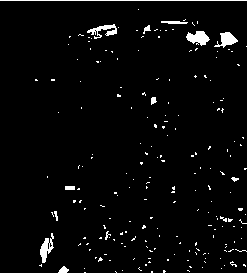} &&&  \includegraphics[width=.30\linewidth]{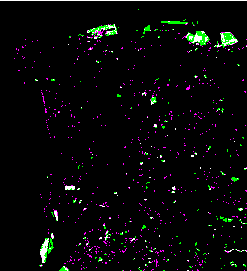} &
    \includegraphics[width=.30\linewidth]{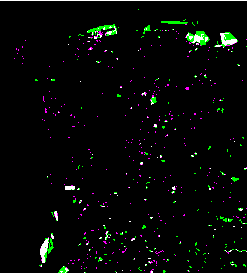}\\
    Ground truth&&& \multicolumn{2}{c}{Linear classifier} \\
    
    \end{tabular}

    \caption{Example of change detection for the proposed method. {True positives are depicted in white, missed changes in green and false positives in magenta.}
    }
    \label{fig:visu}
\end{figure}

The best performance among the self-supervised approaches, top half of Tab.~\ref{tab:SOTA_change_detection}, was achieved by the model pretrained on pretext Task 2 combined with the {CVA} classifier using the triangle method. This leads to the highest F1-score. The CVA with the Otsu method has the highest sensitivity (recall, meaning that the most changes are detected), but at the price of a very low precision due to the very high number of false positives; see also the maps in Fig.~\ref{fig:visu}. This is most probably due to the setting of the Otsu threshold, which needs to be very high to favor sensitivity. The learned classifiers (`linear') in Table~\ref{tab:SOTA_change_detection} provide the best results for pretext Task 1 and also the best results in both tasks in terms of specificity, but also show lower sensitivity scores. This results in a slightly lower F1-score for pretext Task 2. Compared with current state of art in the OSCD dataset, the self supervised models perform remarkably well, given its shallow architecture and the fact that they are pre-trained in an unsupervised way.

\begin{table}[!t]
\small
\caption[]{Comparison between supervised State of Art (S-o-A) and Self supervised models on the undisclosed test set of OCSD. All metrics are expressed in percentage. The best performance as measured by each metric are highlighted in bold text. 'Linear' corresponds to a learned linear classifier for change detection.}  

\centering
\begin{tabular}{lp{2.5cm}ccccc} \hline
&Method && Sensitivity & Specificity & Precision & F1 \\
\hline\hline
\multirow{6}{*}{\rotatebox{90}{Self-supervised}}
&Task 1 & CVA+Otsu & 65.78 & 86.18 & 20.60 & 31.37 \\
&Task 1 & CVA+Triangle & 41.14 & {96.11} & 36.55 &  38.71 \\
&Task 1 & linear & 50.00  & {96.66} & 37.98 & 43.17 \\

&Task 2 & CVA+Otsu & \textbf{83.85} & 81.99 & 20.24  & 32.61 \\
&Task 2 & CVA+Triangle & 52.80 & 95.76 & {40.42}  & \textbf{45.79} \\
&Task 2 & linear & 35.37 & \textbf{97.76} & \textbf{46.30} & 43.17 \\
\hline\hline
\multirow{5}{*}{\rotatebox{90}{S-o-A}}
&Siamese & \cite{Daudt:2018:OSCD_CD_dataset}  & \textbf{85.63} & 85.35 & 24.16  & 37.69 \\
&Early fusion& \cite{Daudt:2018:OSCD_CD_dataset} & 84.69 & 88.33 & 28.34  & 42.47 \\
&FC-EF& \cite{Daudt:2018:fully_convolutional} & 50.97 & \textbf{98.51} & \textbf{64.42} &  56.91 \\
&FC-Siam-Conv& \cite{Daudt:2018:fully_convolutional} & 65.15 & 95.23 & 42.39  & 51.36 \\
&FC-Siam-Diff& \cite{Daudt:2018:fully_convolutional} & 57.99 & 97.73 & 57.81  & \textbf{57.91} \\
\hline
\end{tabular}
\label{tab:SOTA_change_detection}
\end{table}

Finally, Fig.~\ref{fig:visu} illustrates some change maps for the Beirut image of the OCSD dataset. Looking at the maps, we observe that the CVA detection is accurate on the top right corner, but also that it tends to generate more false positives (in {magenta}), and, when using the Otsu method, {most of the} image is predicted as changed. We therefore conclude that Otsu's method is inferior to the other two, which can be both considered usable. Remarkably, the learned classifier reduces the false positive and shows the most visually pleasant results, but at the price of less precise delineation of the change than CVA with the triangle method.

\section{Conclusions}
In this paper, we explored the possibility of pre-training a convolutional neural network for change detection without labels. We perform such training by forging a pretext task inherent in the data, which aims at learning a feature space where unchanged pixels are close and far from abnormal situations. We use two self-supervised learning approaches and then fine tune the network trained this way to detect changes. Experiments in the benchmark Sentinel-2 OCSD dataset shows that traininig a model this way can lead to results close to state of the art deep learning change detection. {It is available at the URL \url{https://zenodo.org/record/4280482}}.

\bibliographystyle{ieeetr}
\bibliography{myBibliography}

\begin{thebibliography}{10}

\bibitem{Liu:2019:review_multispectral_cd}
S.~Liu, D.~Marinelli, L.~Bruzzone, and F.~Bovolo, ``{A review of change
  detection in multitemporal hyperspectral images: Current techniques,
  applications, and challenges},'' {\em IEEE Geoscience and Remote Sensing
  Magazine}, vol.~7, no.~2, pp.~140--158, 2019.

\bibitem{Verbesselt:2010:trend_seasonal_timeseries}
J.~Verbesselt, R.~Hyndman, G.~Newnham, and D.~Culvenor, ``Detecting trend and
  seasonal changes in satellite image time series,'' {\em Remote Sensing of
  Environment}, vol.~114, no.~1, pp.~106--115, 2010.

\bibitem{Hamunyela:2016:spatial_context_bfast}
E.~Hamunyela, J.~Verbesselt, and M.~Herold, ``{Using spatial context to improve
  early detection of deforestation from Landsat time series},'' {\em Remote
  Sensing of Environment}, vol.~172, pp.~126--138, 2016.

\bibitem{Deng:2009:urbanization}
J.~S. Deng, K.~Wang, Y.~Hong, and J.~G. Qi, ``{Spatio-temporal dynamics and
  evolution of land use change and landscape pattern in response to rapid
  urbanization},'' {\em Landscape and Urban Planning}, vol.~92, no.~3-4,
  pp.~187--198, 2009.

\bibitem{Huang:2017:urbanisation}
X.~Huang, D.~Wen, J.~Li, and R.~Qin, ``{Multi-level monitoring of subtle urban
  changes for the megacities of China using high-resolution multi-view
  satellite imagery},'' {\em Remote Sensing of Environment}, vol.~196,
  pp.~56--75, 2017.

\bibitem{Brunner:2010:multi_sensor_CD_disaster}
D.~Brunner, G.~Lemoine, and L.~Bruzzone, ``{Earthquake damage assessment of
  buildings using VHR optical and SAR imagery},'' {\em IEEE Transactions on
  Geoscience and Remote Sensing}, vol.~48, no.~5, pp.~2403--2420, 2010.

\bibitem{Longbotham:2012:data_fusion_contest_cd}
N.~Longbotham, F.~Pacifici, T.~Glenn, A.~Zare, M.~Volpi, D.~Tuia,
  E.~Christophe, J.~Michel, J.~Inglada, J.~Chanussot, and Q.~Du, ``{Multi-modal
  change detection, application to the detection of flooded areas: Outcome of
  the 2009-2010 data fusion contest},'' {\em IEEE Journal of Selected Topics in
  Applied Earth Observations and Remote Sensing}, vol.~5, no.~1, pp.~331--342,
  2012.

\bibitem{Bovolo:2015:time_cd}
F.~Bovolo and L.~Bruzzone, ``{The Time Variable in Data Fusion: A Change
  Detection Perspective},'' {\em IEEE Geoscience and Remote Sensing Magazine},
  vol.~3, no.~3, pp.~8--26, 2015.

\bibitem{Vol10e}
M.~Volpi, D.~{Tuia}, F.~Bovolo, M.~Kanevski, and L.~Bruzzone, ``Supervised
  change detection in {VHR} images using contextual information and support
  vector machines,'' {\em Int. J. Appl. Earth Obs. Geoinf.}, vol.~20,
  pp.~77--85, 2013.

\bibitem{Vol14b}
M.~Volpi, G.~Camps-Valls, and D.~{Tuia}, ``Spectral alignment of cross-sensor
  images with automated kernel canonical correlation analysis,'' {\em ISPRS J.
  Int. Soc. Photo. Remote Sens.}, vol.~107, pp.~50--63, 2015.

\bibitem{Lin:2019:multispectral_bilinearCNN}
Y.~Lin, S.~Li, L.~Fang, and P.~Ghamisi, ``Multispectral change detection with
  bilinear convolutional neural networks,'' {\em IEEE Geoscience and Remote
  Sensing Letters}, 2019.

\bibitem{Zhan:2017:siamese_cd}
Y.~Zhan, K.~Fu, M.~Yan, X.~Sun, H.~Wang, and X.~Qiu, ``Change detection based
  on deep siamese convolutional network for optical aerial images,'' {\em IEEE
  Geoscience and Remote Sensing Letters}, vol.~14, no.~10, pp.~1845--1849,
  2017.

\bibitem{Peng:2019:UNet++}
D.~Peng, Y.~Zhang, and H.~Guan, ``End-to-end change detection for high
  resolution satellite images using improved unet++,'' {\em Remote Sensing},
  vol.~11, no.~11, 2019.

\bibitem{Mou:2019:CNN+RNN}
L.~Mou, L.~Bruzzone, and X.~X. Zhu, ``{Learning spectral-spatialoral features
  via a recurrent convolutional neural network for change detection in
  multispectral imagery},'' {\em IEEE Transactions on Geoscience and Remote
  Sensing}, vol.~57, no.~2, pp.~924--935, 2019.

\bibitem{Saha:2019:deepCVA}
S.~Saha, F.~Bovolo, and L.~Bruzzone, ``Unsupervised deep change vector analysis
  for multiple-change detection in vhr images,'' {\em IEEE Transactions on
  Geoscience and Remote Sensing}, vol.~57, no.~6, pp.~3677--3693, 2019.

\bibitem{Zhu:2017:DL_remote_sensing}
X.~X. Zhu, D.~Tuia, L.~Mou, G.~S. Xia, L.~Zhang, F.~Xu, and F.~Fraundorfer,
  ``{Deep Learning in Remote Sensing: A Comprehensive Review and List of
  Resources},'' {\em IEEE Geoscience and Remote Sensing Magazine}, vol.~5,
  no.~4, pp.~8--36, 2017.

\bibitem{Kel19d}
B.~Kellenberger, D.~Marcos, and D.~{Tuia}, ``When a few clicks make all the
  difference: Improving weakly-supervised wildlife detection in {UAV} images,''
  in {\em IEEE/CVF Conference on Computer Vision and Pattern Recognition
  Workshops (CVPRW)}, (Long Beach, CA), 2019.

\bibitem{daudt2019gad}
R.~{Caye Daudt}, B.~{Le Saux}, A.~Boulch, and Y.~Gousseau, ``Guided anisotropic
  diffusion and iterative learning for weakly supervised change detection,'' in
  {\em Computer Vision and Pattern Recognition Workshops}, June 2019.

\bibitem{Doersch:2015:self-supervised_spatial_context}
C.~Doersch, A.~Gupta, and A.~A. Efros, ``Unsupervised visual representation
  learning by context prediction,'' in {\em IEEE International Conference on
  Computer Vision (ICCV)}, pp.~1422--1430, 2015.

\bibitem{Caron:2018:self-supervised_clustering}
M.~Caron, P.~Bojanowski, A.~Joulin, and M.~Douze, ``Deep clustering for
  unsupervised learning of visual features,'' in {\em European Conference on
  Computer Vision (ECCV)}, 2018.

\bibitem{Daudt:2018:OSCD_CD_dataset}
R.~C. {Daudt}, B.~{Le Saux}, A.~{Boulch}, and Y.~{Gousseau}, ``Urban change
  detection for multispectral earth observation using convolutional neural
  networks,'' in {\em IEEE International Geoscience and Remote Sensing
  Symposium (IGARSS)}, pp.~2115--2118, 2018.

\bibitem{Gidaris:2018:self-supervised_rotation}
S.~Gidaris, P.~Singh, and N.~Komodakis, ``Unsupervised representation learning
  by predicting image rotations,'' in {\em International Conference on Learning
  Representations (ICLR)}, 2018.

\bibitem{dataset_simplemaps}
simplemaps, ``{World Cities Database},'' 2019.

\bibitem{dataset_geonames}
geonames, ``{Major cities of the world},'' 2019.

\bibitem{sentinel:2015:user_handbook}
Suhet, ``Sentinel-2 user handbook,'' 2015.

\bibitem{kingma:2014:adam}
D.~P. Kingma and J.~Ba, ``Adam: {A} method for stochastic optimization,'' in
  {\em International Conference on Learning Representations (ICLR)}, 2015.

\bibitem{Bruzzone:2013:cd_framework}
L.~Bruzzone and F.~Bovolo, ``{A novel framework for the design of
  change-detection systems for very-high-resolution remote sensing images},''
  {\em Proceedings of the IEEE}, vol.~101, no.~3, pp.~609--630, 2013.

\bibitem{Rosin:2001:triangle}
P.~L. Rosin, ``Unimodal thresholding,'' {\em Pattern Recognition}, vol.~34,
  no.~11, pp.~2083 -- 2096, 2001.

\bibitem{Yosinski:2014:transferring_features}
J.~Yosinski, J.~Clune, Y.~Bengio, and H.~Lipson, ``How transferable are
  features in deep neural networks?,'' in {\em Advances in Neural Information
  Processing Systems (NIPS)}, pp.~3320--3328, 2014.

\bibitem{Kolesnikov:2019:self_supervised_comparison}
A.~Kolesnikov, X.~Zhai, and L.~Beyer, ``Revisiting self-supervised visual
  representation learning,'' in {\em IEEE Conference on Computer Vision and
  Pattern Recognition (CVPR)}, 2019.

\bibitem{Daudt:2018:fully_convolutional}
R.~C. {Daudt}, B.~{Le Saux}, and A.~{Boulch}, ``Fully convolutional siamese
  networks for change detection,'' in {\em IEEE International Conference on
  Image Processing (ICIP)}, pp.~4063--4067, 2018.

\end{thebibliography}

\end{document}